\documentclass{article}

\PassOptionsToPackage{numbers, compress}{natbib}

\usepackage[preprint]{nips_2018}

\usepackage[utf8]{inputenc} 
\usepackage[T1]{fontenc}    
\usepackage{hyperref}       
\usepackage{url}            
\usepackage{booktabs}       
\usepackage{amsfonts}       
\usepackage{nicefrac}       
\usepackage{microtype}      
\usepackage{graphicx}
\usepackage{algorithm}
\usepackage{algpseudocode}
\usepackage{amsmath,amssymb} 
\usepackage{xcolor}
\usepackage{multirow}

\title{Image Inpainting via Generative Multi-column Convolutional Neural Networks}

\author{
    Yi Wang$^{1}$ \quad Xin Tao$^{1,2}$ \quad Xiaojuan Qi$^{1}$ \quad Xiaoyong Shen$^{2}$ \quad Jiaya Jia$^{1,2}$\\
    $^{1}$The Chinese University of Hong Kong \quad $^{2}$YouTu Lab, Tencent\\
    \texttt{\{yiwang, xtao, xjqi, leojia\}@cse.cuhk.edu.hk \quad goodshenxy@gmail.com} 
}

\newcommand{\bs}{\mathbf{s}}
\newcommand{\bv}{\mathbf{v}}
\newcommand{\br}{\mathbf{r}}

\newcommand{\X}{\mathbf{X}}
\newcommand{\Y}{\mathbf{Y}}
\newcommand{\M}{\mathbf{M}}
\newcommand{\Mw}{\mathbf{M}_w}

\newcommand{\flY}{\Phi_{L}({\mathbf{Y}})}

\newcommand{\eg}{{\textit{e.g.}}}
\newcommand{\etal}{{\textit{et al.}}}

\begin{document}

    \maketitle

\begin{abstract}
In this paper, we propose a generative multi-column network for image inpainting. This network synthesizes different image components in a parallel manner within one stage. To better characterize global structures, we design a confidence-driven reconstruction loss while an implicit diversified MRF regularization is adopted to enhance local details. The multi-column network combined with the reconstruction and MRF loss propagates local and global information derived from context to the target inpainting regions. Extensive experiments on challenging street view, face, natural objects and scenes manifest that our method produces visual compelling results even without previously common post-processing.
\end{abstract}

\section{Introduction}

Image inpainting (also known as image completion) aims to estimate suitable pixel
information to fill holes in images. It serves various applications such as object
removal, image restoration, image denoising, to name a few. Though studied for many
years, it remains an open and challenging problem since it is highly ill-posed. In order
to generate realistic structures and textures, researchers resort to auxiliary
information, from either surrounding image areas or external data.

A typical inpainting method exploits pixels under certain patch-wise similarity measures,
addressing three important problems respectively to (1) extract suitable features to
evaluate patch similarity; (2) find neighboring patches; and (3) to aggregate auxiliary
information.

\begin{figure}
        \centering
        \includegraphics[width=1\linewidth]{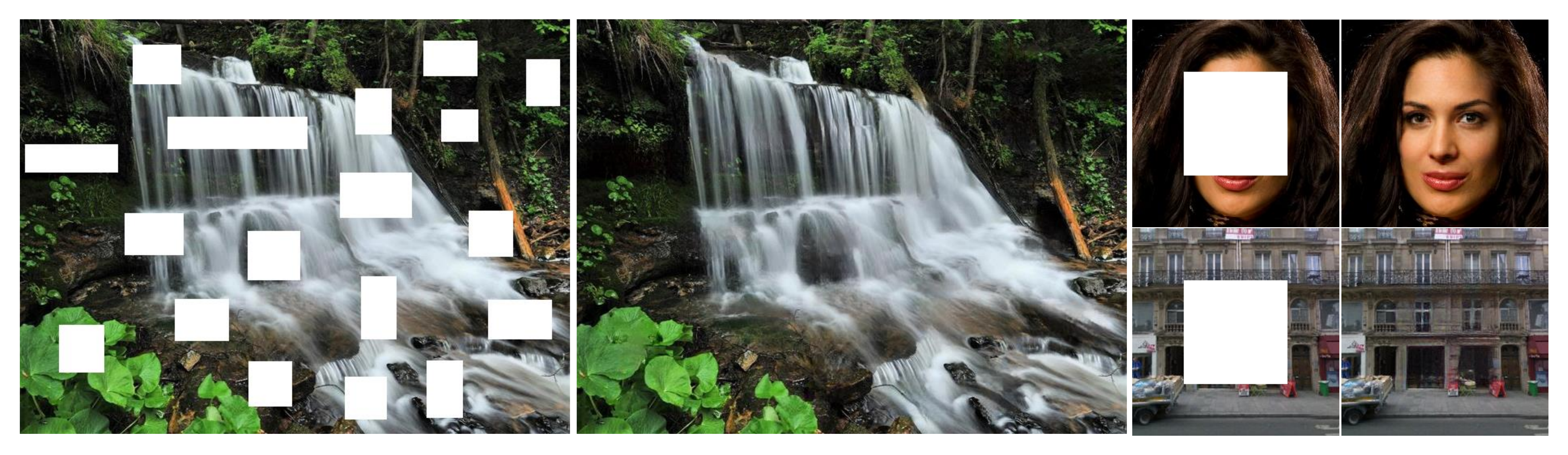}\vspace{-0.2in}
        \caption{Our inpainting results on building, face, and natural scene.}\vspace{-0.1in}
        \label{fig:teaser}
    \end{figure}

\paragraph{Features for Inpainting}
Suitable feature representations are very important to build connections between missing
and known areas. In contrast to traditional patch-based methods using hand-crafted
features, recent learning-based algorithms learn from data. From the model perspective,
inpainting requires understanding of global information. For example, only by seeing the
entire face, the system can determine eyes and nose position, as shown in top-right of
Figure~\ref{fig:teaser}. On the other hand, pixel-level details are crucial for visual
realism, \eg~texture of the skin/facade in Figure~\ref{fig:teaser}.

Recent CNN-based methods utilize encoder-decoder
\cite{pathak2016context,yeh2017semantic,yang2017high,iizuka2017globally,yu2018generative}
networks to extract features and achieve impressive results. But there is still much room
to consider features as a group of different components and combine both global semantics
and local textures.

\paragraph{Reliable Similar Patches}
In both exemplar-based
\cite{he2012statistics,he2014image,criminisi2004region,sun2005image,jia2003image,jia2004inference,barnes2009patchmatch}
and recent learning-based methods
\cite{pathak2016context,yeh2017semantic,yang2017high,iizuka2017globally,yu2018generative},
explicit nearest-neighbor search is one of the key components for generation of realistic
details. When missing areas originally contain structure different from context, the
found neighbors may harm the generation process. Also, nearest-neighbor search during
testing is also time-consuming. Unlike these solutions, we in this paper apply search
only in the training phase with improved similarity measure. Testing is very efficient
without the need of post-processing.

\paragraph{Spatial-variant Constraints}
Another important issue is that inpainting can take multiple candidates to fill holes.
Thus, optimal results should be constrained in a spatially variant way -- pixels close to
area boundary are with few choices, while the central part can be less constrained. In
fact, adversarial loss has already been used in recent methods
\cite{pathak2016context,yeh2017semantic,yang2017high,iizuka2017globally,yu2018generative}
to learn multi-modality. Various weights are applied to loss
\cite{pathak2016context,yeh2017semantic,yu2018generative} for boundary consistency. In
this paper, we design a new spatial-variant weight to better handle this issue.

The overall framework is a \emph{Generative Multi-column Convolutional Neural Network}
(GMCNN) for image inpainting. The multi-column
structure~\cite{ciregan2012multi,zhang2016single,agostinelli2013adaptive} is used since
it can decompose images into components with different receptive fields and feature
resolutions. Unlike multi-scale or coarse-to-fine strategies
\cite{yang2017high,karras2017progressive} that use resized images, branches in our
multi-column network directly use full-resolution input to characterize multi-scale
feature representations regarding global and local information. A new \emph{implicit
diversified Markov random field} (ID-MRF) term is proposed and used in the training phase
only. Rather than directly using the matched feature, which may lead to visual artifacts,
we incorporate this term as regularization.

Additionally, we design a new \emph{confidence-driven reconstruction loss} that
constrains the generated content according to the spatial location. With all these
improvements, the proposed method can produce high quality results considering boundary
consistency, structure suitability and texture similarity, without any post-processing
operations. Exemplar inpainting results are given in Figure \ref{fig:teaser}.

\section{Related Work}
\textbf{Exemplar-based Inpainting~~} Among traditional methods, exemplar-based inpainting
\cite{he2012statistics,he2014image,criminisi2004region,sun2005image,jia2003image,jia2004inference,barnes2009patchmatch}
copies and pastes matching patches in a pre-defined order. To preserve structure, patch
priority computation specifies the patch filling order
\cite{criminisi2004region,he2012statistics,he2014image,sun2005image}. With only low-level
information, these methods cannot produce high-quality semantic structures that do not
exist in examples, {\eg}, faces and facades.

    \textbf{CNN Inpainting}
    Since the seminal \textit{context-encoder} work \cite{pathak2016context}, deep CNNs have achieved significant progress. Pathak {\etal} proposed training an encoder-decoder CNN and minimizing pixel-wise reconstruction loss and adversarial loss. Built upon \textit{context-encoder}, in \cite{iizuka2017globally}, global and local discriminators helped improve the adversarial loss where a fully convolutional encoder-decoder structure was adopted. Besides encoder-decoder, U-net-like structure was also used \cite{yan2018shift}.

    Yang \etal \cite{yang2017high} and Yu \etal \cite{yu2018generative} introduced coarse-to-fine CNNs for image inpainting. To generate more plausible and detailed texture, combination of CNN and Markov Random Field \cite{yang2017high} was taken as the post-process to improve inpainting results from the coarse CNN. It is inevitably slow due to iterative MRF inference. Lately, Yu {\etal} conducted nearest neighbor search in deep feature space \cite{yu2018generative}, which brings clearer texture to the filling regions compared with previous strategies of a single forward pass.

    \section{Our Method}
    Our inpainting system is trainable in an end-to-end fashion, which takes an image $\mathbf{X}$ and a binary region mask $\mathbf{M}$ (with value 0 for known pixels and 1 otherwise) as input. Unknown regions in image $\mathbf{X}$ are filled with zeros. It outputs a complete image $\mathbf{\hat{Y}}$. We detail our network design below.

    \subsection{Network Structure}

    \begin{figure}[t]
        \centering
        \includegraphics[width=0.95\linewidth]{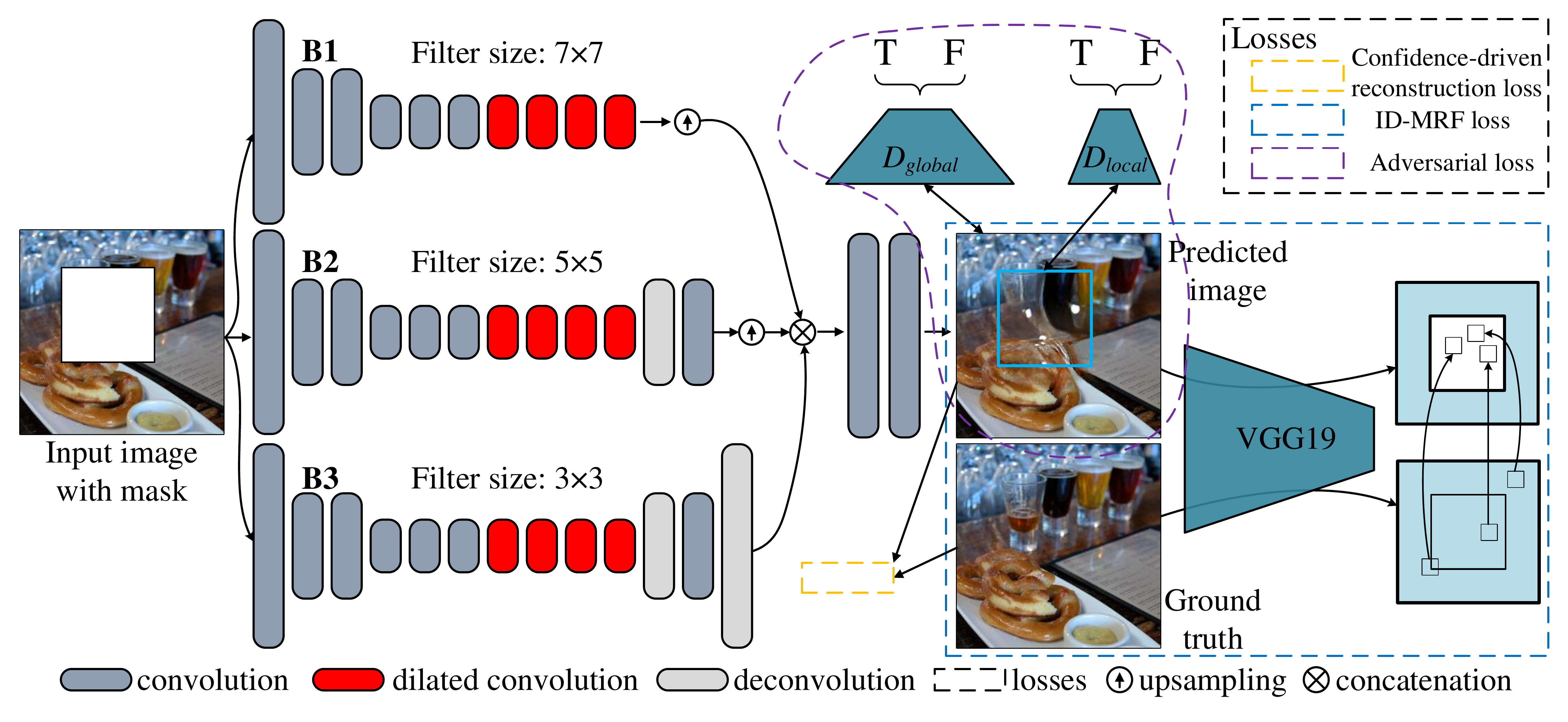}\vspace{-0.2in}
        \caption{Our framework.}
        \label{fig:netarchitecture}
    \end{figure}

    Our proposed \emph{Generative Multi-column Convolutional Neural Network} (GMCNN) shown in Figure~\ref{fig:netarchitecture} consists of three sub-networks: a generator to produce results, global\&local discriminators for adversarial training, and a pretrained VGG network \cite{simonyan2014very} to calculate ID-MRF loss. In the testing phase, only the generator network is used.

    The generator network consists of $n$ ($n=3$) parallel encoder-decoder branches to extract different levels of features from input ${\X}$ with mask ${\M}$, and a shared decoder module to transform deep features into natural image space $\mathbf{\hat{Y}}$.
    We choose various receptive fields and spatial resolutions for these branches as shown in Figure~\ref{fig:netarchitecture}, which capture different levels of information.
    Branches are denoted as $\{f_{i}(\cdot)\}$ ($i\in\{1,2,...,n\}$), trained in a data driven manner to generate better feature components than handcrafted decomposition.

    Then these components are up-sampled (bilinearly) to the original resolution and are concatenated into feature map $F$. We further transform features $F$ into image space via shared decoding module with 2 convolutional layers, denoted as $d(\cdot)$. The output is $\mathbf{\hat{Y}}=d(F)$. Minimizing the difference between $\mathbf{\hat{Y}}$ and ${\Y}$ makes $\{f_{i}(\cdot)\}_{i=1,...,n}$ capture appropriate components in ${\X}$ for inpainting. $d(\cdot)$ further transforms such deep features to our desired result. Note that although $f_i(\cdot)$ seems independent of each other, they are mutually influenced during training due to $d(\cdot)$.

    \paragraph{Analysis} Our framework is by nature different from commonly used one-stream encoder-decoder structure and the coarse-to-fine architecture~\cite{yang2017high,yu2018generative,karras2017progressive}. The encoder-decoder transforms the image into a common feature space with the same-size receptive field, ignoring the fact that inpainting involves different levels of representations. The multi-branch encoders in our GMCNN contrarily do not have this problem. Our method also overcomes the limitation of the coarse-to-fine architecture, which paints the missing pixels from small to larger scales where errors in the coarse-level already influence refinement. Our GMCNN incorporates different structures in parallel. They complement each other instead of simply inheriting information.

    \begin{figure}[t]
    \centering
    \begin{tabular}{ccc}
        \multicolumn{3}{c}{\includegraphics[width=0.95\linewidth]{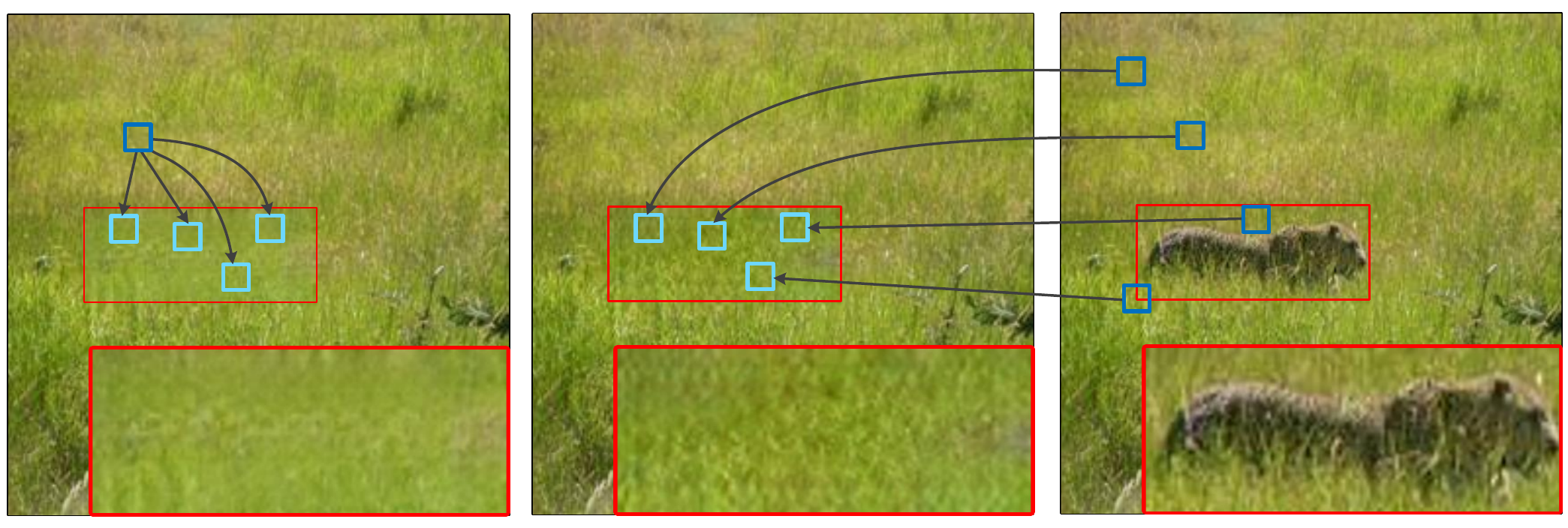}}\\
        \hspace{0.14\columnwidth}(a) & \hspace{0.26\columnwidth}(b) & \hspace{0.14\columnwidth}(c)\\
    \end{tabular}
    \caption{Using different similarity measures to search the nearest neighbors. (a) Inpainting results using cosine similarity. (b) Inpainting results using our relative similarity. (c) Ground truth image where red rectangle highlights the filling region (Best viewed in original resolution and with color).}
    \label{fig:mrfdrawback}
\end{figure}

    \subsection{ID-MRF Regularization}
    Here, we address aforementioned semantic structure matching and computational-heavy iterative MRF optimization issues. Our scheme is to take MRF-like regularization only in the training phase, named {\it implicit diversified Markov random fields} (ID-MRF). The proposed network is optimized to minimize the difference between generated content and corresponding nearest-neighbors from ground truth in the feature space. Since we only use it in training, complete ground truth images make it possible to know high-quality nearest neighbors and give appropriate constraints for the network.

    To calculate ID-MRF loss, it is possible to simply use direct similarity measure ({\eg} cosine similarity) to find the nearest neighbors for patches in generated content. But this procedure tends to yield smooth structure, as a flat region easily connects to similar patterns and quickly reduces structure variety, as shown in Figure \ref{fig:mrfdrawback}(a). We instead adopt a relative distance measure \cite{mechrez2018contextual,mechrez2018learning,talmi2017template} to model the relation between local features and target feature set. It can restore subtle details as illustrated in Figure \ref{fig:mrfdrawback}(b).

    Specifically, let $\mathbf{\hat{Y}}_g$ be the generated content for the missing regions, $\mathbf{\hat{Y}}_g^L$ and ${\Y}^L$ are the features generated by the $L_{th}$ feature layer of a pretrained deep model. For neural patches ${\bv}$ and ${\bs}$ extracted from $\mathbf{\hat{Y}}_g^L$ and ${\Y}^L$ respectively, the relative similarity from ${\bv}$ to ${\bs}$ is defined as
    \begin{equation}
    {\rm RS}({\bv},{\bs})=\exp((\frac{\mu({\bv},{\bs})}{\max_{{\br} \in \rho_{\bv}({\Y}^L)}\mu({\bv},{\br})+\epsilon})/h),
    \end{equation}
    where $\mu(\cdot,\cdot)$ is the cosine similarity. ${\br} \in \rho_{\bv}({\Y}^L)$ means ${\br}$ belongs to ${\Y}^L$ excluding ${\bv}$. $h$ and $\epsilon$ are two positive constants. If ${\bv}$ is like ${\bs}$ more than other neural patches in ${\flY}$, ${\rm RS}({\bv},{\bs})$ turns large.

    Next, ${\rm RS}({\bv},{\bs})$ is normalized as
    \begin{equation} \label{eq_rd_}
    \overline{\rm RS}({\bv},{\bs})={\rm RS}({\bv},{\bs})/\sum_{{\br} \in \rho_{\bv}({\Y}^L)}{\rm RS}({\bv},{\br}).
    \end{equation}
    Finally, with Eq. \eqref{eq_rd_}, the ID-MRF loss between ${\mathbf{\hat{Y}}_g^L}$ and ${\Y}^L$ is defined as
    \begin{equation}
    \mathcal{L}_{M}(L) = -\log(\frac{1}{Z}\sum_{{\bs} \in {{\Y}^L}}\max_{{\bv} \in {\mathbf{\hat{Y}}_g^L}}\overline{{\rm RS}}({\bv},{\bs})),
    \end{equation}
    where $Z$ is a normalization factor. For each ${\bs} \in {{\Y}^L}$, ${\bv bet}=\arg \max_{{\bv} \in {\mathbf{\hat{Y}}_g^L}} \overline{{\rm RS}}({\bv},{\bs})$ means ${\bv}$ is closer to ${\bs}$ compared with other neural patches in ${\mathbf{\hat{Y}}_g^L}$. In the extreme case that all neural patches in ${\mathbf{\hat{Y}}_g^L}$ are close to one patch ${\bs}$, other patches ${\br}$ have their $\max_{{\bv}} \overline{{\rm RS}}({\bv},{\br})$ small. So $\mathcal{L}_{M}(L)$ is large.

    On the other hand, when the patches in ${\mathbf{\hat{Y}}_g^L}$ are close to different candidates in ${\Y}^L$, each ${\br}$ in ${\Y}^L$ has its unique nearest neighbor in ${\mathbf{\hat{Y}}_g^L}$. The resulting $\max_{{\bv} \in {\mathbf{\hat{Y}}_g^L}} \overline{{\rm RS}}({\bv},{\br})$ is thus big and $\mathcal{L}_{M}(L)$ becomes small. We show one example in the supplementary file. From this perspective, minimizing $\mathcal{L}_{M}(L)$ encourages each ${\bv}$ in ${\mathbf{\hat{Y}}_g^L}$ to approach different neural patches in ${\Y}^L$, diversifying neighbors, as shown in Figure \ref{fig:mrfdrawback}(b).

    An obvious benefit for this measure is to improve the similarity between feature distributions in ${\mathbf{\hat{Y}}_g^L}$ and ${\Y}^L$. By minimizing the ID-MRF loss, not only local neural patches in ${\mathbf{\hat{Y}}_g^L}$ find corresponding candidates from ${\Y}^L$, but also the feature distributions come near, helping capture variation in complicated texture.

    Our final ID-MRF loss is computed on several feature layers from VGG19. Following common practice \cite{gatys2016image,li2016combining}, we use \textit{conv4\_2} to describe image semantic structures. Then \textit{conv3\_2} and \textit{conv4\_2} are utilized to describe image texture as
    \begin{equation}
    \mathcal{L}_{mrf} = \mathcal{L}_{M}(conv4\_2) + \sum_{\mathbf{t}=3}^{4}\mathcal{L}_{M}(conv\mathbf{t}\_2).
    \end{equation}

    \paragraph{More Analysis} During training, ID-MRF regularizes the generated content based on the reference. It has the strong ability to create realistic texture locally and globally. We note the fundamental difference from the methods of \cite{yang2017high,yu2018generative}, where nearest-neighbor search via networks is employed in the testing phase. Our ID-MRF regularization exploits both reference and contextual information inside and out of the filling regions, and thus causes high diversity in inpainting structure generation.

    \subsection{Information Fusion}

    \paragraph{Spatial Variant Reconstruction Loss} Pixel-wise reconstruction loss is important for inpainting \cite{pathak2016context,yeh2017semantic,yu2018generative}. To exert constraints based on spatial location, we design the confidence-driven reconstruction loss where unknown pixels close to the filling boundary are more strongly constrained than those away from it. We set the confidence of known pixels as 1 and unknown ones related to the distance to the boundary. To propagate the confidence of known pixels to unknown ones, we use a Gaussian filter $g$ to convolve $\overline{\M}$ to create a loss weight mask ${\Mw}$ as
    \begin{equation} \label{eq:reconstruction}
    {\mathbf{M}_w^{i}} = (g * \overline{\M}^{i}) \odot {\M},
    \end{equation}
    where $g$ is with size $64 \times 64$ and its standard deviation is $40$. $\overline{\M}^{i}=\mathbf{1}-{\M}+{\mathbf{M}_w^{i-1}}$ and ${\mathbf{M}_w^{0}}=\mathbf{0}$. $\odot$ is the Hadamard product operator. Eq. \eqref{eq:reconstruction} is repeated several times to generate ${\Mw}$. The final reconstruction loss is
    \begin{equation}
    \mathcal{L}_c=||({\Y}-G([{\X}, {\M}]; \theta))\odot{\Mw}||_1,
    \end{equation}
    where $G([{\X}, {\M}]; \theta)$ is the output of our generative model $G$, and $\theta$ denotes learn-able parameters.

Compared with the reconstruction loss used in \cite{pathak2016context,yeh2017semantic,yu2018generative}, ours exploits spatial locations and their relative order by considering confidence on both known and unknown pixels. It results in the effect of gradually shifting learning focus from filling border to the center and smoothing the learning curve.

    \paragraph{Adversarial Loss}
    Adversarial loss is a catalyst in filling missing regions and becomes common in many creation tasks. Similar to those of \cite{iizuka2017globally,yu2018generative}, we apply the improved Wasserstein GAN \cite{gulrajani2017improved} and use local and global discriminators. For the generator, the adversarial loss is defined as
    \begin{equation}
    \mathcal{L}_{adv}=-E_{{\X} \sim \mathbb{P}_{\X}}[D(G({\X}; \theta))]+\lambda_{gp} E_{\hat{{\X}}\sim \mathbb{P}_{\hat{{\X}}}}[(||\nabla_{\hat{{\X}}}D(\hat{{\X}})\odot {\Mw}||_2-1)^2],
    \end{equation}
    where $\hat{{\X}}=tG([{\X}, {\M}];\theta)+(1-t){\Y}$ and $t \in [0, 1]$.

    \subsection{Final Objective}
    With confidence-driven reconstruction loss, ID-MRF loss, and adversarial loss, the model objective of our net is defined as
    \begin{equation}
    \mathcal{L}=\mathcal{L}_{c}+\lambda_{mrf} \mathcal{L}_{mrf}+\lambda_{adv} \mathcal{L}_{adv},
    \end{equation}
    where $\lambda_{adv}$ and $\lambda_{mrf}$ are used to balance the effects between local structure regularization and adversarial training.

    \subsection{Training}
    We train our model first with only confidence-driven reconstruction loss and set $\lambda_{mrf}$ and $\lambda_{adv}$ to 0s to stabilize the later adversarial training. After our model $G$ converges, we set $\lambda_{mrf}=0.05$ and $\lambda_{adv}=0.001$ for fine tuning until converge. The training procedure is optimized using Adam solver \cite{kingma2014adam} with learning rate $1e-4$. We set $\beta_1=0.5$ and $\beta_2=0.9$. The batch size is 16.

    For an input image ${\Y}$, a binary image mask ${\M}$ (with value 0 for known and 1 for unknown pixels) is sampled at a random location. The input image ${\X}$ is produced as ${\X}={\Y}\odot(\mathbf{1}-{\M})$. Our model $G$ takes the concatenation of ${\X}$ and ${\M}$ as input. The final prediction is $\hat{{\Y}}={\Y} \odot (\mathbf{1}-{\M}) +G([{\X},{\M}]) \odot {\M}$. All input and output are linearly scaled within range $[-1,1]$.

    \section{Experiments}
    We evaluate our method on five datasets of Paris street view \cite{pathak2016context}, Places2 \cite{zhou2017places}, ImageNet \cite{russakovsky2015imagenet}, CelebA \cite{liu2015deep}, and CelebA-HQ \cite{karras2017progressive}.

    \subsection{Experimental Settings}
    We train our models on the training set and evaluate our model on the testing set (for Paris street view) or validation set (for Places2, ImageNet, CelebA, and CelebA-HQ). In training, we use images of resolution $256\times256$ with the largest hole size $128\times128$ in random positions. For Paris street view, places2, and ImageNet, $256 \times 256$ images are randomly cropped and scaled from the full-resolution images. For CelebA and CelebA-HQ face datasets, images are scaled to $256 \times 256$. All results given in this paper are not post-processed.

    Our implementation is with Tensorflow v1.4.1, CUDNN v6.0, and CUDA v8.0. The hardware is with an Intel CPU E5 (2.60GHz) and TITAN X GPU. Our model costs 49.37ms and 146.11ms per image on GPU for testing images with size $256\times256$ and $512\times512$, respectively. Using ID-MRF in training phrase costs 784ms more per batch (with 16 images of $256 \times 256 \times 3$ pixels). The total number of parameters of our generator network is 12.562M.

    \subsection{Qualitative Evaluation}
\begin{figure}
        \begin{center}
            \centering
            \begin{tabular}{ccccc}
                \multicolumn{5}{c}{\includegraphics[width=0.95\linewidth]{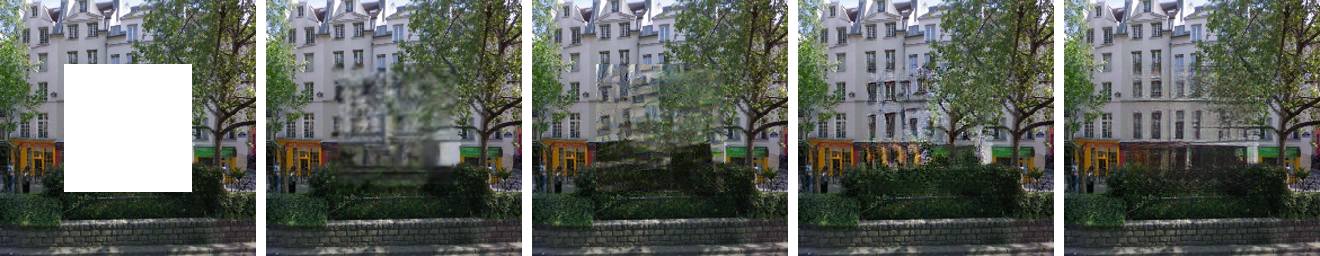}}\\
                \multicolumn{5}{c}{\includegraphics[width=0.95\linewidth]{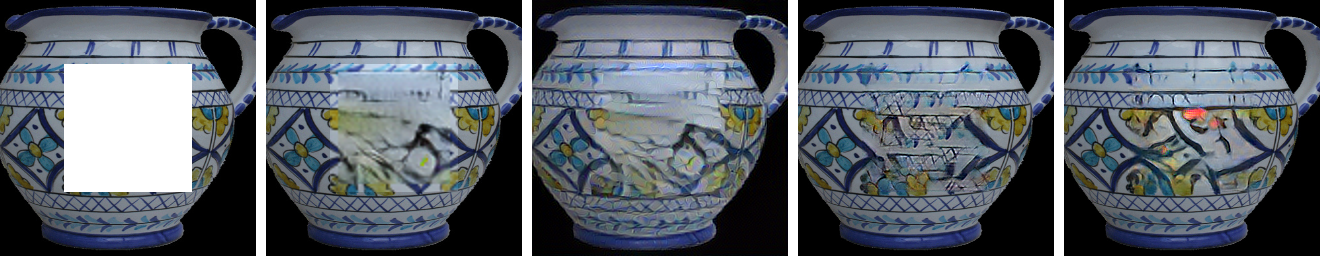}}\\
                \hspace{0.08\columnwidth}(a) & \hspace{0.13\columnwidth}(b) & \hspace{0.13\columnwidth}(c) & \hspace{0.13\columnwidth}(d) & \hspace{0.06\columnwidth}(e)\\
            \end{tabular}
        \end{center}
        \caption{Visual comparisons on Paris street view (up) and ImageNet (down). (a) Input image. (b) CE \cite{pathak2016context}. (c) MSNPS \cite{yang2017high}. (d) CA \cite{yu2018generative}. (e) Our results (best viewed in higher resolution).}
        \label{fig_paris_results}
    \end{figure}

\begin{figure}
        \begin{center}
            \centering
            \begin{tabular}{ccc}
                \multicolumn{3}{c}{\includegraphics[width=1\linewidth]{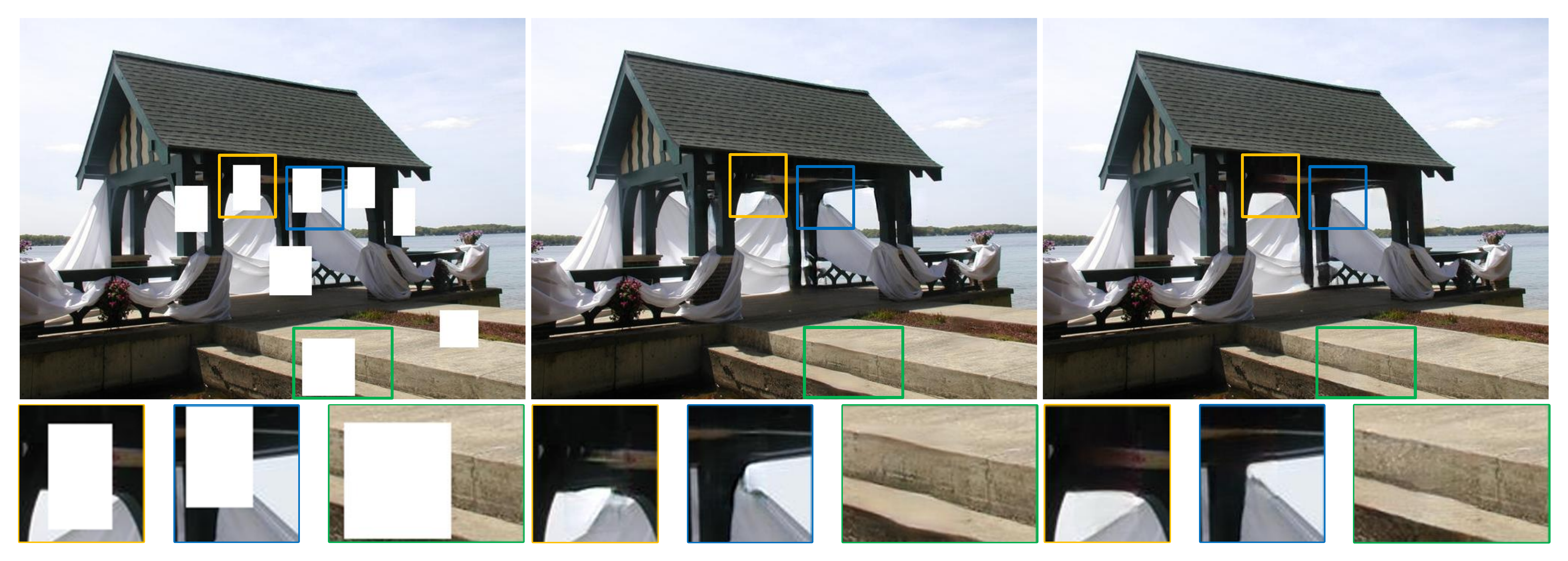}}\\
                \hspace{0.16\columnwidth}(a) & \hspace{0.27\columnwidth}(b) & \hspace{0.14\columnwidth}(c)\\
            \end{tabular}
        \end{center}
        \caption{Visual comparisons on Places2 for $512 \times 680$ images with random masks. (a) Input image. (b) Results by CA \cite{yu2018generative}. (c) Our results.}
        \label{fig_placeshd}
    \end{figure}

    As shown in Figures \ref{fig_paris_results} and \ref{fig_placeshd}, compared with other methods, ours gives obvious visual improvement on plausible image structures and crisp textures. The more reasonably generated structures mainly stem from the multi-column architecture and confidence-driven reconstruction loss. The realistic textures are created via ID-MRF regularization and adversarial training by leveraging the contextual and corresponding textures.

\begin{figure}
        \begin{center}
            \centering
            \begin{tabular}{cccccc}
                \multicolumn{3}{c}{\includegraphics[width=0.48\linewidth]{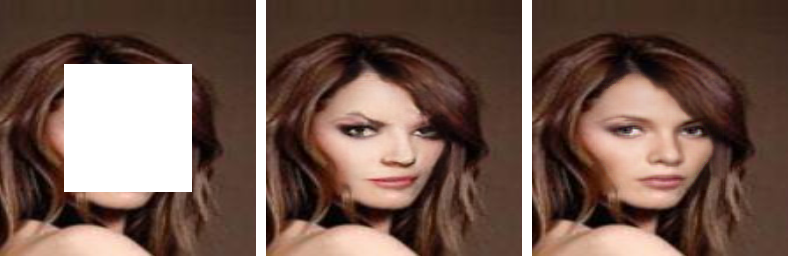}} &
                \multicolumn{3}{c}{\includegraphics[width=0.48\linewidth]{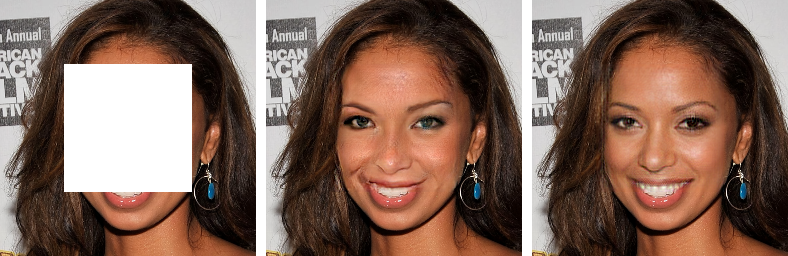}}\\
                \hspace{0.07\columnwidth}(a) & \hspace{0.10\columnwidth}(b) & \hspace{0.05\columnwidth}(c) & \hspace{0.07\columnwidth}(a) & \hspace{0.10\columnwidth}(b) & \hspace{0.05\columnwidth}(c)\\
            \end{tabular}
        \end{center}
        \caption{Visual comparisons on CelebA (Left) and CelebA-HQ (Right). (a) Input image. (b) CA \cite{yu2018generative}. (c) Our results. }
        \label{fig_celeba_results}
    \end{figure}

    In Figures \ref{fig_celeba_results}, we show partial results of our method and CA \cite{yu2018generative} on CelebA and CelebA-HQ face datasets. Since we do not apply MRF in a non-parametric manner, visual artifacts are much reduced. It is notable that finding suitable patches for these faces is challenging. Our ID-MRF regularization remedies the problem. Even the face shadow and reflectance can be generated as shown in Figure \ref{fig_celeba_results}.

    Also, our model is trained with arbitrary-location and -size square masks. It is thus general to be applied to different-shape completion as shown in Figures \ref{fig_placeshd} and \ref{fig:teaser}. More inpainting results are in our project website.

    \subsection{Quantitative Evaluation}
    Although the generation task is not suitable to be evaluated by peak signal-to-noise ratio (PSNR) or structural similarity (SSIM), for completeness, we still give them on the testing or validation sets of four used datasets for reference. In ImageNet, only 200 images are randomly chosen for evaluation since MSNPS \cite{yang2017high} takes minutes to complete a $256 \times 256$ size image. As shown in Table \ref{tb_index}, our method produces decent results with comparable or better PSNR and SSIM.

    We also conduct user studies as shown in Table~\ref{tb_user_study}. The protocol is based on large batches of blind randomized A/B tests deployed on the Google Forms platform. Each survey involves a batch of 40 pairwise comparisons. Each pair contains two images completed from the same corrupted input by two different methods. There are 40 participants invited for user study. The participants are asked to select the more realistic image in each pair. The images are all shown at the same resolution ($256 \times 256$).
    The comparisons are randomized across conditions and the left-right order is randomized. All images are shown for unlimited time and the participant is free to spend as much time as desired on each pair. In all conditions, our method outperforms the baselines.

    \begin{table}
        \centering
        \small
        \begin{tabular}{c|cccccccc}
            \hline
            \multirow{2}*{Method} & \multicolumn{2}{c}{Pairs street view-100} & \multicolumn{2}{c}{ImageNet-200} & \multicolumn{2}{c}{Places2-2K} &  \multicolumn{2}{c}{CelebA-HQ-2K}\\
            ~ & PSNR & SSIM & PSNR & SSIM & PSNR & SSIM & PSNR & SSIM\\
            \hline
            CE [18] &
            $23.49$ & $\mathbf{0.8732}$ &
            $\mathbf{23.56}$ & $\mathbf{0.9105}$ &
            $-$ & $-$ &
            $-$ & $-$ \\
            MSNPS [24] &
            $24.44$ & $0.8477$ &
            $20.62$ & $0.7217$ &
            $-$ & $-$ &
            $-$ & $-$ \\
            CA [26] &
            $23.78$ & $0.8588$ &
            $22.44$ & $0.8917$ &
            $20.03$ & $0.8539$ &
            $23.98$ & $0.9441$\\
            Ours &
            $\mathbf{24.65}$ & $0.8650$ &
            $22.43$ & $0.8939$ &
            $\mathbf{20.16}$ & $\mathbf{0.8617}$ &
            $\mathbf{25.70}$ & $\mathbf{0.9546}$\\
            \hline
        \end{tabular}
        \caption{Quantitative results on five testing datasets.}
        \vspace{-0.2in}
        \label{tb_index}
    \end{table}

    \begin{table}
        \centering
        \small
        \begin{tabular}{c|ccccc}
            \hline
            ~ & Paris street view & ImageNet & Places2 & CelebA & CelebA-HQ \\
            \hline
            GMCNN > CE [18] & $98.1\%$ & $88.3\%$ & - & - & - \\
            GMCNN > MSNPS [24] & $94.4\%$ & $86.5\%$ & - & - & - \\
            GMCNN > CA [26] & $84.2\%$ & $78.5\%$ & $69.6\%$ & $99.0\%$ & $93.8\%$ \\
            \hline
        \end{tabular}
        \caption{Result of user study. Each entry is the percentage of cases where results by our approach are judged more realistic than another solution.}
        \vspace{-0.2in}
        \label{tb_user_study}
    \end{table}

\begin{table}
    \centering
    \small
    \begin{tabular}{c|ccccc}
        \hline
        Model & Encoder-decoder& Coarse-to-fine & GMCNN-f & GMCNN-v w/o ID-MRF & GMCNN-v \\
        \hline
        PSNR & 23.75 & 23.63 & 24.36 & 24.62 & $\mathbf{24.65}$\\
        SSIM & 0.8580 & 0.8597 & 0.8644 & $\mathbf{0.8657}$ & 0.8650\\
        \hline
    \end{tabular}
    \caption{Quantitative results of different structures on Paris street view dataset (-f/-v: fixed/varied receptive fields).}
    \vspace{-0.2in}
    \label{tb_index_ab}
\end{table}

\begin{table}[!h]
    \centering
    \small
    \begin{tabular}{c|cccc}
        \hline
        $\lambda_{mrf}$ & 2 & 0.2 & 0.02 & 0.002\\
        \hline
        PSNR & 24.62 & 24.53 & $\mathbf{24.64}$ & 24.36\\
        SSIM & $\mathbf{0.8659}$ & 0.8652 & 0.8654 & 0.8640\\
        \hline
    \end{tabular}
    \caption{Quantitative results about how ID-MRF regularizes the inpainting performance.}
    \vspace{-0.2in}
    \label{tb_change_id_mrf}
\end{table}

    \subsection{Ablation Study}

\begin{figure}[!h]
        \begin{center}
            \centering
            \begin{tabular}{ccccc}
                \multicolumn{5}{c}{\includegraphics[width=0.95\linewidth]{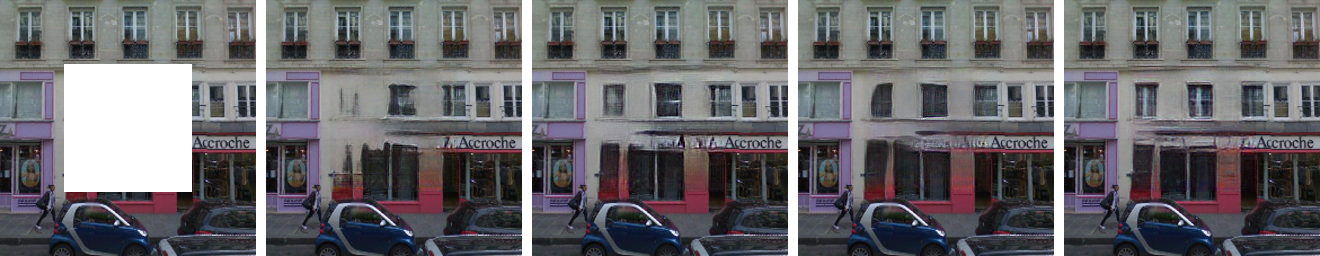}}\\
                \hspace{0.08\columnwidth}(a) & \hspace{0.13\columnwidth}(b) & \hspace{0.13\columnwidth}(c) & \hspace{0.13\columnwidth}(d) & \hspace{0.06\columnwidth}(e)\\
            \end{tabular}
        \end{center}
        \caption{Visual comparison of CNNs with different structures. (a) Input image. (b) Single encoder-decoder. (c) Coarse-to-fine structure \cite{yu2018generative}. (d) GMCNN with the fixed receptive field in all branches. (e) GMCNN with varied receptive fields.}
        \label{fig_ablation_1}
    \end{figure}

    \textbf{Single Encoder-Decoder \textit{vs.} Coarse-to-Fine \textit{vs.} GMCNN~~} We evaluate our multi-column architecture by comparing with single encode-decoder and coarse-to-fine networks with two sequential encoder-decoder (same as that in \cite{yu2018generative} except no contextual layer). The single encoder-decoder is just the same as our branch three (\textbf{B3}). To minimize the influence of model capacity, we triple the filter sizes in the single encoder-decoder architecture to make its parameter size as close to ours as possible. The loss for these three structures is the same, including confidence-driven reconstruction loss, ID-MRF loss, and WGAN-GP adversarial loss. The corresponding hyper-parameters are the same. The testing results are shown in Figure \ref{fig_ablation_1}. Our GMCNN structure with varied receptive fields in each branch predicts reasonable image structure and texture compared with single encoder-decoder and coarse-to-fine structure. Additional quantitative experiment is given in Table \ref{tb_index_ab}, showing the proposed structure is beneficial to restore image fidelity.

    \textbf{Varied Receptive Fields \textit{vs.} Fixed Receptive Field~~}
    We then validate the necessity of using varied receptive fields in branches. The GMCNN with the same receptive field in each branch turns to using 3 identical third Branches in Figure \ref{fig:netarchitecture} with filter size $5 \times 5$. Figure \ref{fig_ablation_1} shows within the GMCNN structure, branches with varied receptive fields give visual more appealing results.

\begin{figure}[!h]
        \begin{center}
            \centering
            \begin{tabular}{cccccc}
                \multicolumn{3}{c}{\includegraphics[width=0.48\linewidth]{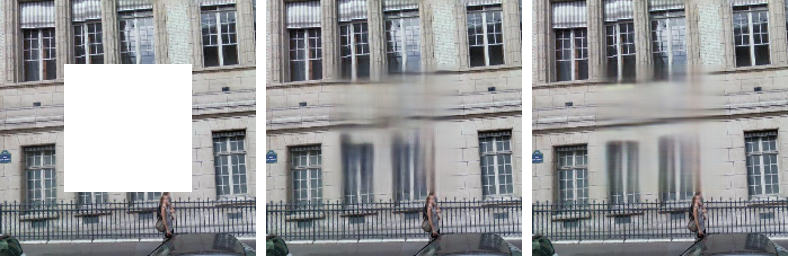}} & \multicolumn{3}{c}{\includegraphics[width=0.48\linewidth]{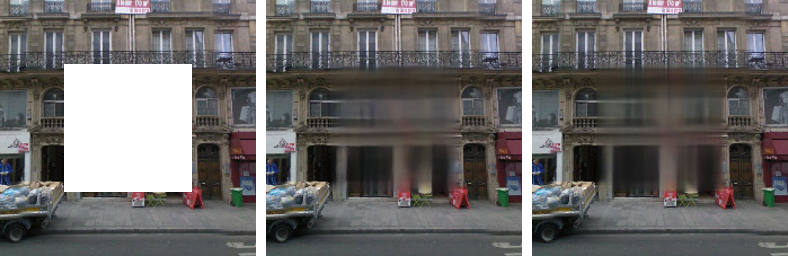}}\\
                \hspace{0.07\columnwidth}(a) & \hspace{0.10\columnwidth}(b) & \hspace{0.05\columnwidth}(c) & \hspace{0.07\columnwidth}(a) & \hspace{0.10\columnwidth}(b) & \hspace{0.05\columnwidth}(c)\\
            \end{tabular}
        \end{center}
        \caption{Visual comparisons of different reconstruction losses. (a) Input image. (b) Spatial discounted loss \cite{yu2018generative}. (c) Confidence-driven reconstruction loss.}
        \label{fig_ablation_2}
    \end{figure}

    \textbf{Spatial Discounted Reconstruction Loss \textit{vs.} Confidence-Driven Reconstruction Loss~~} We compare our confidence-driven reconstruction loss with alternative spatial discounted reconstruction loss \cite{yu2018generative}. We use a single-column CNN trained only with the losses on the Paris street view dataset. The testing results are given in Figure \ref{fig_ablation_2}. Our confidence-driven reconstruction loss works better.

    \begin{figure}[!h]
    \begin{center}
        \centering
        \begin{tabular}{cccccc}
            \multicolumn{3}{c}{\includegraphics[width=0.48\linewidth]{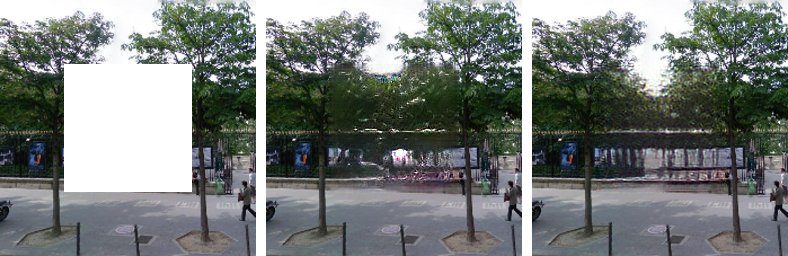}} & \multicolumn{3}{c}{\includegraphics[width=0.48\linewidth]{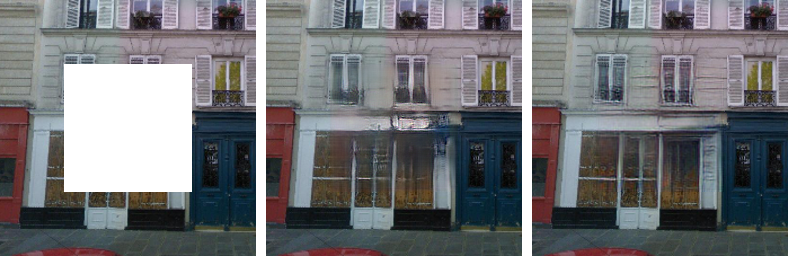}}\\
            \hspace{0.07\columnwidth}(a) & \hspace{0.10\columnwidth}(b) & \hspace{0.05\columnwidth}(c) & \hspace{0.07\columnwidth}(a) & \hspace{0.10\columnwidth}(b) & \hspace{0.05\columnwidth}(c)\\
        \end{tabular}
    \end{center}
    \caption{Visual comparison of results using ID-MRF and not with it. (a) Input image. (b) Results using ID-MRF. (c) Results without using ID-MRF.}
    \label{fig_ablation_3}
    \end{figure}

\begin{figure}[!h]
    \begin{center}
        \centering
        \begin{tabular}{ccccc}
            \multicolumn{5}{c}{\includegraphics[width=0.95\linewidth]{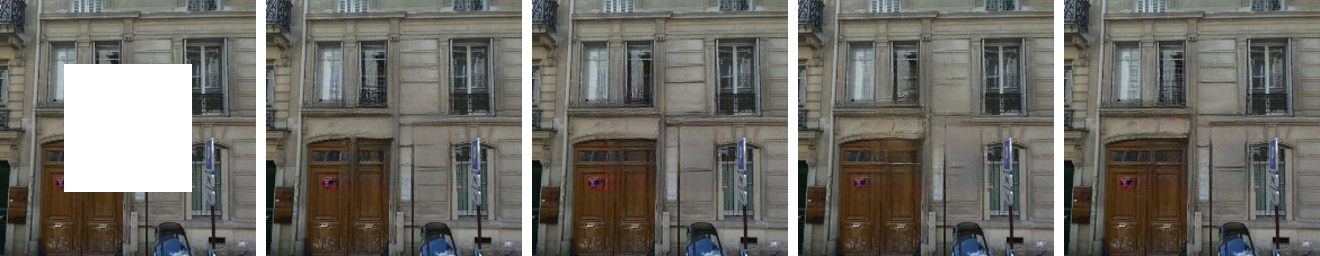}}\\
            \hspace{0.08\columnwidth}(a) & \hspace{0.13\columnwidth}(b) & \hspace{0.13\columnwidth}(c) & \hspace{0.13\columnwidth}(d) & \hspace{0.06\columnwidth}(e)\\
        \end{tabular}
    \end{center}
    \caption{Visual comparison of results using ID-MRF with different $\lambda_{mrf}$ (a) Input image. (b) $\lambda_{mrf}=2$. (c) $\lambda_{mrf}=0.2$. (d) $\lambda_{mrf}=0.02$. (e) $\lambda_{mrf}=0.002$.}
    \label{fig_ablation_change_idmrf}
\end{figure}

    \textbf{With and without ID-MRF Regularization~~} We train a complete GMCNN on the Paris street view dataset with all losses and one model that does not involve ID-MRF. As shown in Figure \ref{fig_ablation_3}, ID-MRF can significantly enhance local details. Also, the qualitative and quantitative changes are given in Table \ref{tb_change_id_mrf} and Figure \ref{fig_ablation_change_idmrf} about how $\lambda_{mrf}$ affects inpainting performance. Empirically, $\lambda_{mrf}=0.02\sim0.05$ strikes a good balance.

    \section{Conclusion}
    We have primarily addressed the important problems of representing visual context and using it to generate and constrain unknown regions in inpainting. We have proposed a generative multi-column neural network for this task and showed its ability to model different image components and extract multi-level features. Additionally, the ID-MRF regularization is very helpful to model realistic texture with a new similarity measure. Our confidence-driven reconstruction loss also considers spatially variant constraints. Our future work will be to explore other constraints with location and content.
    
    \paragraph{Limitations}
    Similar to other generative neural networks \cite{pathak2016context,yang2017high,yu2018generative,yeh2017semantic} for inpainting, our method still has difficulties dealing with large-scale datasets with thousands of diverse object and scene categories, such as ImageNet. When data falls into a few categories, our method works best, since the ambiguity removal in terms of structure and texture can be achieved in these cases.

    \clearpage

    \bibliographystyle{ieee}
    \bibliography{egbib}

\end{document}